\documentclass[11pt]{article}

\usepackage[final]{acl}

\usepackage{times}
\usepackage{latexsym}
\usepackage[T1]{fontenc}
\usepackage[utf8]{inputenc}
\usepackage{microtype}
\usepackage{inconsolata}

\usepackage{graphicx}
\usepackage{booktabs}
\usepackage{multirow}
\usepackage{amsmath}
\usepackage{amssymb}
\usepackage{tcolorbox}
\usepackage{float}
\usepackage{algorithm}
\usepackage{algpseudocode}
\usepackage{xcolor}
\usepackage{fvextra}

\DefineVerbatimEnvironment{Highlighting}{Verbatim}{
  breaklines=true,
  breakanywhere=true,
  fontsize=\scriptsize
}

\title{One Prompt Does Not Fit All: Self-Meta-Evolve for Personalized Information Extraction}

\author{ \textbf{Hongliang Li\textsuperscript{1,2}\thanks{Work conducted during Hongliang Li's internship at Microsoft.}}, \textbf{Lu Wang\textsuperscript{1}}, \textbf{Yong Xu\textsuperscript{1}}, \textbf{Hanyang Chen\textsuperscript{2}}, \textbf{Zhitao Hou\textsuperscript{1}}, \\ \textbf{Xiaoting Qin\textsuperscript{1}}, \textbf{Song Ge\textsuperscript{1}}, \textbf{Qingwei Lin\textsuperscript{1}}, \textbf{Dongmei Zhang\textsuperscript{1}} \\\\ \textsuperscript{1}Microsoft, \textsuperscript{2}Beijing Jiaotong University }

\begin{document}
\maketitle

\begin{abstract}
Large language models (LLMs) are increasingly deployed for enterprise information extraction (IE), where the same document must be reorganized differently for each user. Existing prompt optimization methods, however, rely on a single prompt optimized against a global objective, which is misaligned with the inherent \textit{user heterogeneity} of real workplaces. We formulate enterprise IE as \textbf{per-user prompt adaptation under interaction feedback} and propose \textbf{Self-Meta-Evolve}, a hierarchical framework that maintains a dedicated prompt for each user and continuously refines it through a dual-loop process: an inner loop that edits structured prompts based on persona-conditioned feedback, and an outer loop that evolves the meta-prompt itself by distilling successful editing patterns. To enable scalable training and evaluation, we release a persona-driven IE benchmark of 292 simulated enterprise users, paired with a reproducible persona-generation pipeline grounded in O*NET occupational taxonomies. On this benchmark, Self-Meta-Evolve achieves a 74.58\% success rate, outperforming the strongest prompt-optimization baseline by 13.56 absolute points, and reaches 52.54\% within only two iterations. A double-blind human study with twenty real professionals further confirms that prompts adapted by our framework win against static baselines in 71\% of pairwise comparisons.
\end{abstract}

\section{Introduction}

Large language models (LLMs) are now a core building block of enterprise data pipelines, transforming unstructured artifacts such as emails, meeting transcripts and project reports into structured representations that support downstream analytics and decision-making~\citep{pang2023guideline,xu2024large,liang2026knowledge}. In production, prompts have become the de-facto interface for controlling these systems: many platforms invoke a closed-source LLM via a single carefully engineered prompt and treat the model as a configurable extractor~\citep{xue2024autore,shuang2025utilizing,fuente-etal-2025-guidex}.

\begin{figure}[t]
\centering
\includegraphics[width=\columnwidth]{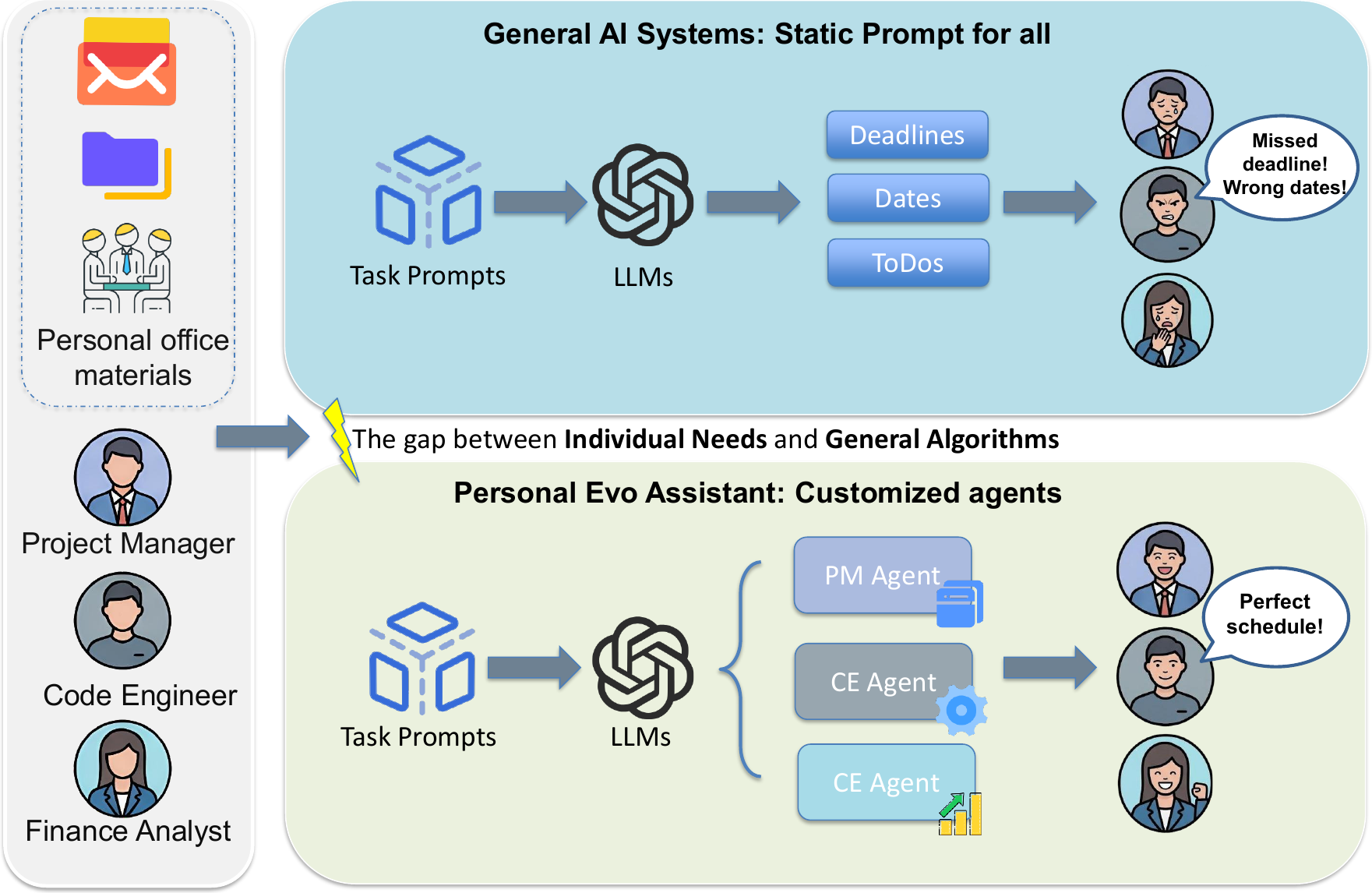}
\caption{Static prompts fail under user-specific extraction objectives. Conventional systems apply a shared prompt to all users, often producing outputs that do not match heterogeneous role-specific preferences. Our framework maintains a dedicated prompt for each user and updates it from interaction feedback, enabling personalized extraction over time.}
\label{fig:introduction_pig}
\end{figure}

This prompt-based paradigm implicitly assumes that one prompt can adequately specify the extraction objective for the entire user population. In practice, however, enterprise users are profoundly heterogeneous~\citep{chen2024persona,shi2025personax,xu-etal-2025-personalized}: as illustrated in Figure~\ref{fig:introduction_pig}, a project manager looking at a project status email cares about actionable deadlines and ownership, whereas a financial analyst reading the same email needs granular budget entities and risk indicators. Our pilot survey of 50 internal users found that 72\% routinely post-edit generic LLM outputs to fit their role-specific standards. Recent enterprise-NLP studies corroborate this gap: when LLM agents are evaluated in realistic workplace sandboxes~\citep{vishwakarma-etal-2025-llms} or applied to heterogeneous enterprise data~\citep{choubey-etal-2025-benchmarking}, role-specific output requirements become the dominant source of failure. Organizational research provides a theoretical lens for this gap: the Job Crafting framework~\citep{wrzesniewski2001crafting} and Work Design Theory~\citep{hackman1976motivation,parker2017one} argue that effective tools must align with the dynamic, role-specific demands that employees actively shape---a property that static, global prompts cannot satisfy.

Existing prompt optimization methods, including search-based~\citep{zhou2022large,yang2023large}, gradient-inspired~\citep{pryzant2023automatic}, evolutionary~\citep{xu-etal-2022-gps,guoconnecting,fernandopromptbreeder}, and multi-agent variants~\citep{zhang-etal-2026-mapro}, share the same global assumption: a single prompt is optimized against a fixed dataset and shared objective. This is fundamentally misaligned with user-centric IE, where the ``correct'' extraction is a function of who is reading it~\citep{guan-etal-2025-survey}.

We instead formulate enterprise IE as \textbf{per-user prompt adaptation under interaction feedback}: each user owns a dedicated prompt that evolves over time, driven by user-specific critique signals. Two observations motivate our design. First, modern LLMs exhibit strong role-playing fidelity, enabling scalable simulation of diverse user personas and their feedback~\citep{park2023generative,tseng2024two,xiao2024learning,wang-etal-2024-rolellm,tamoyan-etal-2025-llm}. Second, prompt optimization decomposes naturally into a hierarchy: low-level edits to a task prompt, and a high-level meta-strategy that decides \emph{how} to edit~\citep{deng2022rlprompt,fernandopromptbreeder,zhang-etal-2026-mapro}.

Building on these observations, we propose \textbf{Self-Meta-Evolve}, a hierarchical framework whose \emph{inner loop} performs persona-conditioned edits to a structured extraction prompt, while an \emph{outer loop} evolves the meta-prompt that controls these edits by distilling successful trajectories. To support scalable yet faithful training, we additionally curate a benchmark of 292 enterprise personas, generated through a reproducible pipeline seeded with O*NET occupational descriptors~\citep{park2023generative,wang2023self}, and validated against real professionals.

Our intended setting is persistent multi-user personalization: editing experience accumulated across users can improve subsequent adaptations. A direct prompt edit remains a practical option for a single user with a simple requirement. Once adapted, a prompt can be reused for subsequent documents without repeating optimization; deployment then requires one extraction call.

We summarize our contributions as follows:
\begin{itemize}
    \itemsep0em
    \item \textbf{Problem formulation.} We formalize enterprise IE as per-user prompt adaptation under interaction feedback, departing from the static, global prompt optimization paradigm.
    \item \textbf{Self-Meta-Evolve framework.} We propose a hierarchical framework that combines per-user prompt editing (inner loop) with shared meta-strategy evolution (outer loop), and analyze how each component contributes to convergence speed and cross-domain generalization.
    \item \textbf{Persona-driven benchmark and human validation.} We release a benchmark of 292 reproducible enterprise personas plus two specialized test sets (STEM, Humanities), and validate the simulation through a double-blind study with twenty real professionals (Cohen's $\kappa$=0.71). Self-Meta-Evolve improves success rate by 13.56 absolute points over the strongest baseline and wins 71\% of pairwise comparisons against a static baseline judged by real users.
\end{itemize}

\section{Related Work}

\paragraph{Prompt optimization.}
Search-based methods such as APE~\citep{zhou2022large} and OPRO~\citep{yang2023large} treat LLMs as black-box optimizers that propose and rank candidate prompts; gradient-inspired methods like ProTeGi~\citep{pryzant2023automatic} use textual critiques as pseudo-gradients. Evolutionary variants~\citep{xu-etal-2022-gps,guoconnecting,chen2023evoprompting,fernandopromptbreeder} apply genetic operators over populations of prompts, and Promptbreeder additionally evolves the mutation prompt itself. Recent work explores meta-learned~\citep{choisystem}, Pareto-optimal~\citep{zhao2025pareto}, and multi-agent~\citep{zhang-etal-2026-mapro} prompt search. All of these optimize a \emph{single} prompt against a \emph{global} objective. Our framework departs from this assumption: prompts are user-specific, and the optimization signal is a persona-conditioned critique rather than ground-truth labels.

\paragraph{LLM-based simulation, judging, and self-improvement.}
LLMs have been used to simulate users~\citep{park2023generative,xiao2024learning,tamoyan-etal-2025-llm} and to judge generations on per-instance criteria~\citep{zheng2023judging,liu-etal-2023-g,fu-etal-2024-gptscore,chirkova-etal-2026-llm}. \citet{dou-etal-2025-simulatorarena} systematically study whether LLM user-simulators are reliable proxies for multi-turn evaluation and find moderate but non-trivial fidelity, motivating cross-model judge checks. Self-improving systems use LLMs to mutate code, heuristics, or prompts via reflective evaluation~\citep{ye2024reevo,liu2024large,wang2026self}. Our work differs in that simulation is \emph{the substrate of personalization}, not just an evaluation tool: AI-User feedback drives per-user adaptation, and the meta-prompt is evolved across users rather than within a single optimization run.

\paragraph{Personalized NLP and role-playing agents.}
Personalization has been studied in dialogue~\citep{zhang-etal-2018-personalizing,madotto-etal-2019-personalizing}, classification~\citep{flek-2020-returning}, fine-grained text generation~\citep{alhafni-etal-2024-personalized}, and recently with LaMP-style benchmarks~\citep{salemi-etal-2024-lamp,salemi-zamani-2025-lamp,salemi-etal-2025-expert}. Surveys on persona/role-playing agents~\citep{tseng2024two,chen2024persona,chen-etal-2025-towards-design,xu-etal-2025-personalized,guan-etal-2025-survey} highlight the gap between role-playing fidelity and downstream task utility, and recent benchmarks evaluate personalization in conversational assistants~\citep{zhao-etal-2025-personalens,mok-etal-2025-exploring} and few-shot personalization with mis-aligned responses~\citep{kim-yang-2025-shot}. Role-playing capabilities have been explicitly elicited in RoleLLM~\citep{wang-etal-2024-rolellm}. At the same time, \citet{kim-etal-2025-persona} and \citet{luz-de-araujo-etal-2026-persistent} point out that personas can have both benefits and drawbacks, and their effectiveness may decline during longer interactions. We account for these issues through an outer-loop reset mechanism (Section~\ref{sec:exp}). Our work introduces personalization into enterprise IE, a commercially important yet relatively underexplored setting, and uses occupational taxonomies as the basis for persona construction to support reproducibility.

\paragraph{LLM-based information extraction and synthetic data.}
LLM-driven IE has been explored for relation~\citep{xue2024autore,jiang-etal-2024-genres}, document-level~\citep{liang2026knowledge}, event~\citep{shuang2025utilizing}, and zero-shot multi-agent~\citep{lu-etal-2025-crossagentie} settings; surveys provide broader coverage~\citep{xu2024large}. Synthetic data has emerged as a key enabler for low-resource and zero-shot IE~\citep{fuente-etal-2025-guidex,ziegler2025craft}. Closest in spirit, personalized meeting summarization systems~\citep{chen-etal-2025-meetalk,kirstein-etal-2025-frame} adapt to user corrections, but operate at the document-summary granularity rather than at the prompt-program level we target. These works optimize a single prompt against gold annotations or fixed user histories. We complement this line by studying \emph{user-relative} correctness, which is the dominant regime in enterprise deployments where ``ground truth'' is role-dependent~\citep{vishwakarma-etal-2025-llms,choubey-etal-2025-benchmarking}.
\begin{figure*}[t]
    \centering
    \includegraphics[width=\textwidth]{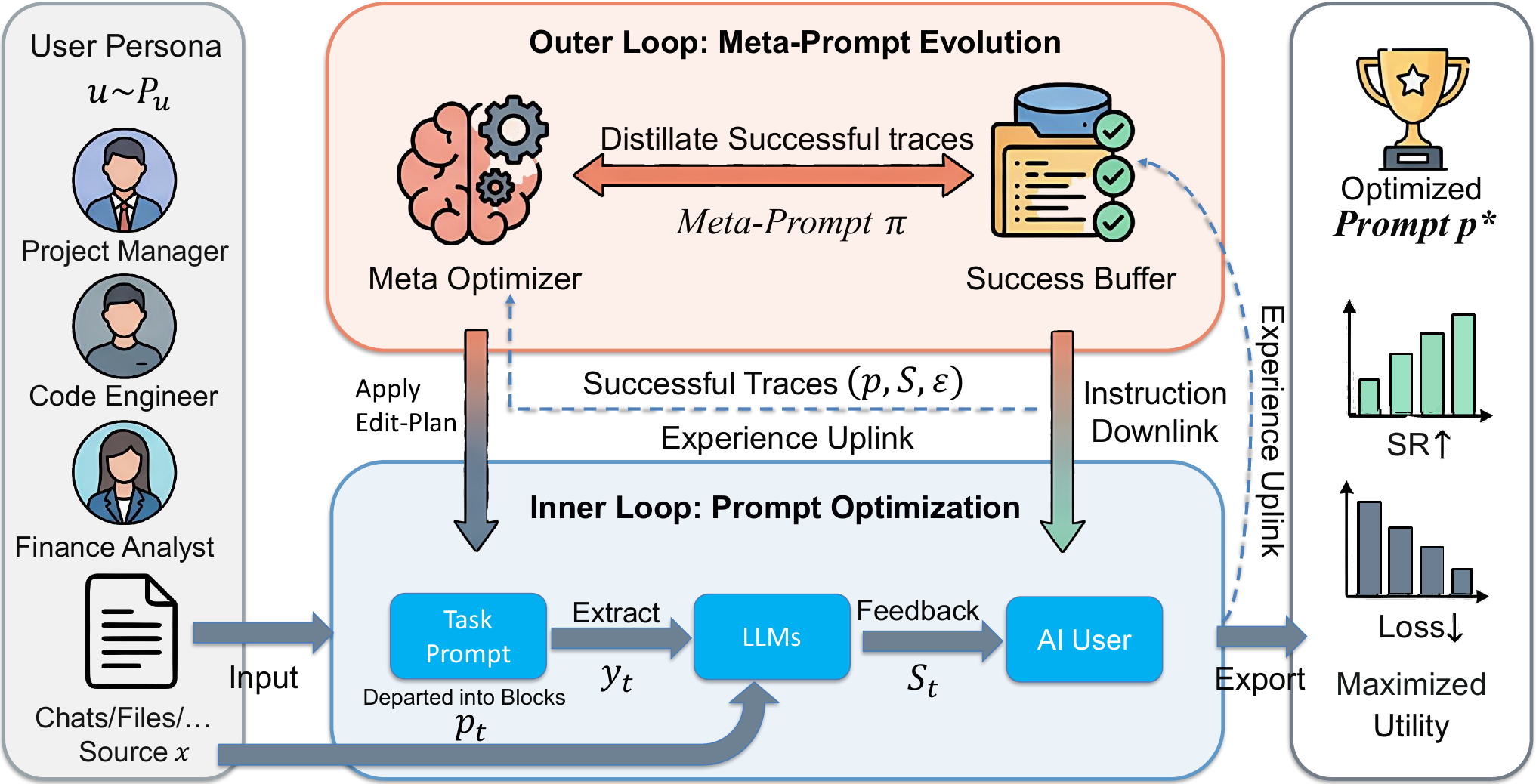}
    \caption{Architecture of Self-Meta-Evolve. The \textbf{inner loop} iteratively refines a user's structured extraction prompt based on AI-User feedback, while the \textbf{outer loop} evolves the meta-prompt by distilling successful editing trajectories across users.}
    \label{fig:framework}
\end{figure*}
\section{Problem Setup}
\label{sec:problem}

\subsection{Per-User Prompt Adaptation}

Let $\mathcal{U}$ denote a distribution over enterprise user personas. Each persona $u \sim P_{\mathcal{U}}$ encodes a role description, a document distribution $\mathcal{D}_u$, and an information preference profile. Given an extraction prompt $p \in \mathcal{P}$ and document $x \sim \mathcal{D}_u$, an LLM produces a structured output $y = f_{\text{LLM}}(x,p)$.

Unlike traditional IE benchmarks that assume static gold annotations, enterprise systems evaluate quality from a \emph{user-centric} perspective: the same $y$ may be acceptable for one user and incorrect for another. We model this by a persona-conditioned feedback function $F_{\phi}(u,x,y)$ that returns structured critiques. The optimization objective is to find a per-user prompt that minimizes the expected feedback-induced loss:
\begin{equation}
p^{*}_u =
\arg\min_{p \in \mathcal{P}}
\mathbb{E}_{x \sim \mathcal{D}_u}
\left[\mathcal{L}(p;u,x)\right],
\end{equation}
\begin{equation}
\mathcal{L}(p;u,x) = g\left(F_{\phi}(u,x,f_{\text{LLM}}(x,p))\right),
\end{equation}
where $g(\cdot)$ converts discrete feedback into a scalar (Section~\ref{sec:ai_user}).

\subsection{AI User Feedback Model}
\label{sec:ai_user}

We instantiate $F_{\phi}$ with an \textbf{AI User} module that role-plays $u$ via LLM prompting and emits structured feedback $F_{\phi}(u,x,y) = \{(a_i, t_i)\}_{i=1}^{m}$, where $a_i$ is an extracted artifact and $t_i \in \{\text{revise}, \text{delete}\}$ specifies the action that the user would take. The scalar loss is
\begin{equation}
\mathcal{L} = \lambda_r N_{\text{revise}} + \lambda_d N_{\text{delete}}.
\end{equation}

\paragraph{Choosing $\lambda_r$ and $\lambda_d$.} The weights reflect the relative cost of \emph{editing} a returned artifact versus \emph{discarding} it entirely. We do not set them arbitrarily: in a pilot annotation study, five enterprise users rated the perceived effort of the two actions across 30 sampled cases on a 1--5 Likert scale, yielding mean ratios of $2.04 \pm 0.41$ in favor of deletion. We accordingly set $\lambda_d = 0.2 = 2\lambda_r$ and verify in Section~\ref{sec:lambda_sens} that final rankings are stable under perturbations of this ratio in $[1.5, 2.5]$.

\paragraph{Loss threshold $\tau$.} A run is \emph{successful} if its deployed loss falls below $\tau$. We pick $\tau$ as the median deployed loss of the strongest baseline (ProTeGi) on a held-out development set of 50 personas, which corresponds to a normalized $\tau = 0.75$. This calibration ensures that ``success'' represents a strict improvement over the prior state of the art, and is reused identically across all methods.

\section{Persona-Driven Enterprise Benchmark}
\label{sec:benchmark}

A central concern for any user-centric IE benchmark is whether the personas faithfully cover real workplace heterogeneity. We therefore describe the construction pipeline in detail (Figure-style overview in Appendix~\ref{app:dataset}) and validate it against real professionals (Section~\ref{sec:human_eval}).

\subsection{Persona Generation Pipeline}

The pipeline is fully reproducible from public seeds and proceeds in four stages.

\paragraph{Stage 1: Seed roles from O*NET.}
We start from the public U.S.\ O*NET occupational database (release 28.0), which provides standardized job titles, task statements, and tools-and-technology lists for $\sim$1{,}000 occupations. We filter to enterprise-relevant occupations using the SOC major groups 11 (Management), 13 (Business and Financial), 15 (Computer and Mathematical), 17 (Engineering), 19 (Life/Physical/Social Sciences), 23 (Legal), 25 (Education), and 27 (Arts/Media). This yields 412 candidate seed roles.

\paragraph{Stage 2: Persona expansion.}
For each seed role, we prompt an LLM (GPT-5.1) to expand the O*NET descriptor into a structured persona JSON with three fields: (i) \emph{role description}, including background, seniority, and primary objectives; (ii) \emph{document distribution}, describing the mix of artifact types such as email, chat, report, and ticket, as well as the relevant topical domains; and (iii) \emph{information preference profile}, which contains a list of 3--6 declarative preferences, for example, ``prioritize tasks with explicit owners and deadlines over status descriptions.'' The expansion prompt and an example showing the transformation from a seed role to a persona are provided in Appendix~\ref{app:persona_prompts}. We generate 400 candidate personas in this stage.

\paragraph{Stage 3: Quality control.}
We apply three filters to obtain the final pool: (a) deduplication via embedding cosine similarity ($>0.92$ collapses to one), removing 47 near-duplicates; (b) consistency check, where an LLM judge scores the persona on internal coherence (1--5) and we drop scores $<4$, removing 38 cases; (c) preference specificity, requiring at least three actionable preferences, removing 23 cases. The remaining \textbf{292 personas} form our main pool. This pipeline transparently explains the 400$\to$292 reduction that earlier iterations of this work were criticized for under-documenting.

\paragraph{Stage 4: Document synthesis.}
For each persona, we synthesize 8--12 enterprise artifacts (emails, chat logs, status reports) conditioned on the persona's document distribution, using a separate generation prompt that injects realistic timestamps, names, and project references. Crucially, the document generator is conditioned only on the \emph{role} and \emph{topical domain}, not on the preference profile, so that documents do not leak the gold preference signal.

For example, a Patent Counsel persona receives fictional invention disclosures, internal patent-review emails, IP-committee meeting notes, and licensing memos, rather than real patents or corporate documents. The generation prompt excludes identifiable names, filing numbers, and jurisdictions.

\subsection{Splits and Test Sets}

We partition the 292 personas by persona ID into 60/20/20 train/dev/eval splits with a fixed seed. To stress cross-domain generalization, we additionally curate two specialized test sets disjoint from the main pool: \textbf{STEM-Personas} ($N=50$, drawn from SOC 15/17/19) and \textbf{Humanities-Personas} ($N=50$, drawn from SOC 25/27 plus librarian/editor occupations). Detailed distributions are in Appendix~\ref{app:dataset}.
\begin{algorithm}[H]
\caption{Self-Meta-Evolve}
\label{alg:meta_prompt_opt}
\begin{algorithmic}[1]
\State \textbf{Init:} prompt $p_0$, meta-prompt $\pi$, buffer $\mathcal{B} \leftarrow \emptyset$
\For{$t = 0$ to $T-1$}
    \State sample persona $u$ and document $x$
    \State $y_t \leftarrow f_{\text{LLM}}(x, p_t)$
    \State $\mathcal{S}_t \leftarrow F_{\phi}(u, x, y_t)$
    \State $E_t \leftarrow f_{\text{LLM}}(p_t, \mathcal{S}_t; \pi)$
    \State $p_{t+1} \leftarrow \text{ApplyEdits}(p_t, E_t)$
    \If{$\mathcal{L}(p_{t+1}) < \mathcal{L}(p_t)$}
        \State $\mathcal{B} \leftarrow \mathcal{B} \cup \{(p_t, \mathcal{S}_t, E_t)\}$
    \EndIf
    \If{$|\mathcal{B}| \bmod K = 0$}
        \State $\pi \leftarrow f_{\text{Meta}}(\pi, \mathcal{B})$
    \EndIf
\EndFor
\end{algorithmic}
\end{algorithm}

\subsection{Data Statement and Release}

All documents are LLM-synthesized; no real enterprise data is used, removing privacy risk. Personas are derived from a public taxonomy. Upon acceptance we release the persona pool, generated documents, AI-User prompt, and pipeline scripts under CC-BY 4.0 to enable replication. Limitations of the synthetic regime are discussed in the Limitations section.

\section{Self-Meta-Evolve Framework}

Our framework (Figure~\ref{fig:framework}) performs per-user prompt adaptation through a bi-level process. The inner loop edits a user's prompt from current-interaction feedback, while the outer loop improves a shared meta-strategy by distilling successful update patterns across users. Pseudo-code is given in~\ref{alg:meta_prompt_opt}.

\subsection{Structured Prompt Representation}

Rather than treating prompts as unstructured text, we represent each extraction prompt as a JSON-like object whose keys are functional sections: task description, extraction instructions, output schema, demonstrations, and constraint rules. Each section is identified by a stable \texttt{point\_id}. This decomposition allows the optimizer to target individual points rather than rewriting the whole prompt, yielding more stable and interpretable updates. The initial prompt is in Appendix~\ref{app:prompts}.

\subsection{Inner Loop: Persona-Conditioned Editing}

At step $t$, a persona $u$ and document $x$ are sampled. The current prompt $p_t$ produces $y_t = f_{\text{LLM}}(x, p_t)$, the AI User returns feedback $\mathcal{S}_t = F_{\phi}(u, x, y_t)$, and an LLM editing agent guided by the meta-prompt $\pi$ proposes an edit plan $E_t = f_{\text{LLM}}(p_t, \mathcal{S}_t; \pi)$ targeting specific (\texttt{section}, \texttt{point\_id}) pairs. The next prompt is
\begin{equation}
p_{t+1} = \text{ApplyEdits}(p_t, E_t).
\end{equation}
If $\mathcal{L}(p_{t+1}; u, x) < \mathcal{L}(p_t; u, x)$, the trace $(p_t, \mathcal{S}_t, E_t)$ is appended to a success buffer $\mathcal{B}$.

\paragraph{Worked example.}
Consider a persona ``DevOps Lead'' whose preference profile demands \emph{actionable on-call follow-ups}. The initial prompt extracts both follow-ups and team announcements; AI-User feedback flags announcements with \texttt{delete}. The inner loop localizes the failure to point \texttt{instr.scope} (``extract all team-relevant updates''), and the editing agent proposes a \texttt{patch\_logic} action that narrows the scope to ``items with explicit owner and deadline.'' Loss drops from 1.4 to 0.6, and the trace enters $\mathcal{B}$.

\subsection{Outer Loop: Meta-Prompt Evolution}

While the inner loop directly improves \emph{prompts}, the outer loop refines the \emph{strategy} for generating edits. Once $|\mathcal{B}|$ reaches a threshold $K$, an LLM acting as a meta-optimizer reads the accumulated successful traces and rewrites the meta-prompt:
\begin{equation}
\pi' = f_{\text{Meta}}(\pi, \mathcal{B}).
\end{equation}

Intuitively, this process captures editing patterns that generalize across users. For example, when feedback indicates over-extraction, the model may prefer applying \texttt{add\_constraint} to \texttt{instr.scope} instead of rewriting demonstrations. These patterns are incorporated into the meta-prompt so that later inner-loop iterations can adapt to \emph{unseen} personas with fewer steps. We therefore assess the quality of the meta-prompt by its adaptation efficiency on held-out personas, which is measured using SR@$t$ (Section~\ref{sec:exp}).

\section{Experiments}
\label{sec:exp}

We address three research questions.
\textbf{RQ1.} Does Self-Meta-Evolve outperform state-of-the-art prompt optimizers on per-user IE?
\textbf{RQ2.} Does the meta-evolutionary mechanism improve early-stage convergence and cross-domain robustness?
\textbf{RQ3.} How does each component contribute, and how does the framework scale with the number of personas?

\subsection{Experimental Setup}

\paragraph{Datasets.} We use the 292-persona main benchmark (60/20/20 split) and the disjoint STEM/Humanities test sets ($N=50$ each).

\paragraph{Baselines.} We compare against: (1) gradient-inspired ProTeGi~\citep{pryzant2023automatic}; (2) search-based APE~\citep{zhou2022large} and OPRO~\citep{yang2023large}; (3) evolutionary EvoPrompt~\citep{chen2023evoprompting}; (4) optimization-specialized Bandit-UCB~\citep{auer2002finite}, MetaSPO~\citep{choisystem}, and Pareto Prompt~\citep{zhao2025pareto}.

\begin{itemize}
    \item \textbf{ProTeGi}~\citep{pryzant2023automatic}: textual gradient via LLM-generated critiques, applied with beam width 4.
    \item \textbf{OPRO}~\citep{yang2023large}: meta-prompt aggregating top-4 historical prompts; 4 candidates generated per step.
    \item \textbf{EvoPrompt}~\citep{chen2023evoprompting}: GA-style evolution with population size 4 and standard NL crossover/mutation.
    \item \textbf{Bandit-UCB}~\citep{auer2002finite}: each prompt section is an arm; UCB1 balances exploration/exploitation.
    \item \textbf{APE}~\citep{zhou2022large}: 10 candidate prompts, dev-set selection.
    \item \textbf{MetaSPO}~\citep{choisystem}: meta-system prompt optimization following the original recipe.
    \item \textbf{Pareto Prompt}~\citep{zhao2025pareto}: dual-objective (accuracy + persona compliance) Pareto front maintenance.
\end{itemize}

\paragraph{Metrics.} \textbf{Success Rate (SR)} is the fraction of personas whose deployed loss is no greater than $\tau$ (Section~\ref{sec:problem}). \textbf{Mean Loss (ML)} averages the deployed loss over personas. \textbf{SR@$t$} is the cumulative success rate within the first $t$ inner-loop iterations and captures convergence speed.

\paragraph{Models.} All methods use the same backbone (GPT-5.1) for both the extractor and the AI User; we additionally validate cross-model robustness in Appendix~\ref{app:cross_model}.

\subsection{Main Results (RQ1)}

Table~\ref{tab:main_results} reports the comparison on the held-out evaluation set. Self-Meta-Evolve achieves \textbf{74.58\% SR}, outperforming the strongest baseline ProTeGi by 13.56 absolute points, and reduces ML from 0.7542 to 0.5122 ($-32\%$). Notably, our method's SR@2 (0.5254) already surpasses the \emph{final} SR of OPRO, EvoPrompt, MetaSPO, APE and Pareto Prompt, confirming that the evolved meta-prompt produces higher-quality edits per step.

A paired bootstrap over matched per-persona success indicators gives $p=0.0038$ against ProTeGi and $p=0.0009$ against Self-Frozen, for SR gains of 13.56 and 16.95 percentage points, respectively.

\begin{table}[t]
\centering
\small
\setlength{\tabcolsep}{3pt}
\begin{tabular}{l|cc|ccc}
\toprule
Method & SR$\uparrow$ & ML$\downarrow$ & SR@2 & SR@4 & SR@8 \\
\midrule
ProTeGi & 0.610 & 0.754 & 0.390 & 0.475 & 0.542 \\
OPRO & 0.458 & 1.124 & 0.271 & 0.339 & 0.424 \\
EvoPrompt & 0.407 & 1.286 & 0.220 & 0.305 & 0.356 \\
Bandit-UCB & 0.542 & 0.941 & 0.322 & 0.424 & 0.492 \\
APE & 0.339 & 1.423 & 0.203 & 0.271 & 0.305 \\
MetaSPO & 0.373 & 1.331 & 0.237 & 0.305 & 0.339 \\
Pareto Prompt & 0.322 & 1.458 & 0.186 & 0.254 & 0.288 \\
\midrule
Self-Frozen (ours) & 0.576 & 0.865 & 0.407 & 0.492 & 0.525 \\
\textbf{Self-Meta-Evolve} & \textbf{0.746} & \textbf{0.512} & \textbf{0.525} & \textbf{0.610} & \textbf{0.678} \\
\bottomrule
\end{tabular}
\caption{Main comparison on the held-out evaluation set ($N=59$). SR@$t$ denotes cumulative success rate at step $t$. Self-Frozen disables the outer loop.}
\label{tab:main_results}
\end{table}

\paragraph{Robustness across backbones.}
Table~\ref{tab:backbone_rebuttal} reports the additional held-out evaluation with five backbones. Self-Meta-Evolve improves over ProTeGi by 11.87--15.25 percentage points in all five cases. This complements the independent-judge checks in Appendix~\ref{app:cross_model}. Additional public-document and shared-objective experiments are reported in Appendix~\ref{app:additional_evaluation}.

\begin{table}[t]
\centering
\small
\setlength{\tabcolsep}{3pt}
\begin{tabular}{lrrr}
\toprule
Backbone & ProTeGi & Ours & $\Delta$ \\
\midrule
GPT-5.5 & 64.41 & 77.97 & 13.56 \\
GPT-5.1 & 61.02 & 74.58 & 13.56 \\
GPT-4.1 & 52.54 & 64.41 & 11.87 \\
Claude-Opus-4.8 & 62.71 & 76.27 & 13.56 \\
Claude-3.7-Sonnet & 54.24 & 69.49 & 15.25 \\
\bottomrule
\end{tabular}
\caption{Additional backbone evaluation: SR (\%) on the held-out set ($N=59$); $\Delta$ is the gain in percentage points.}
\label{tab:backbone_rebuttal}
\end{table}

\subsection{Convergence and Efficiency (RQ2)}

\begin{figure*}[t]
    \centering
    \includegraphics[width=\textwidth]{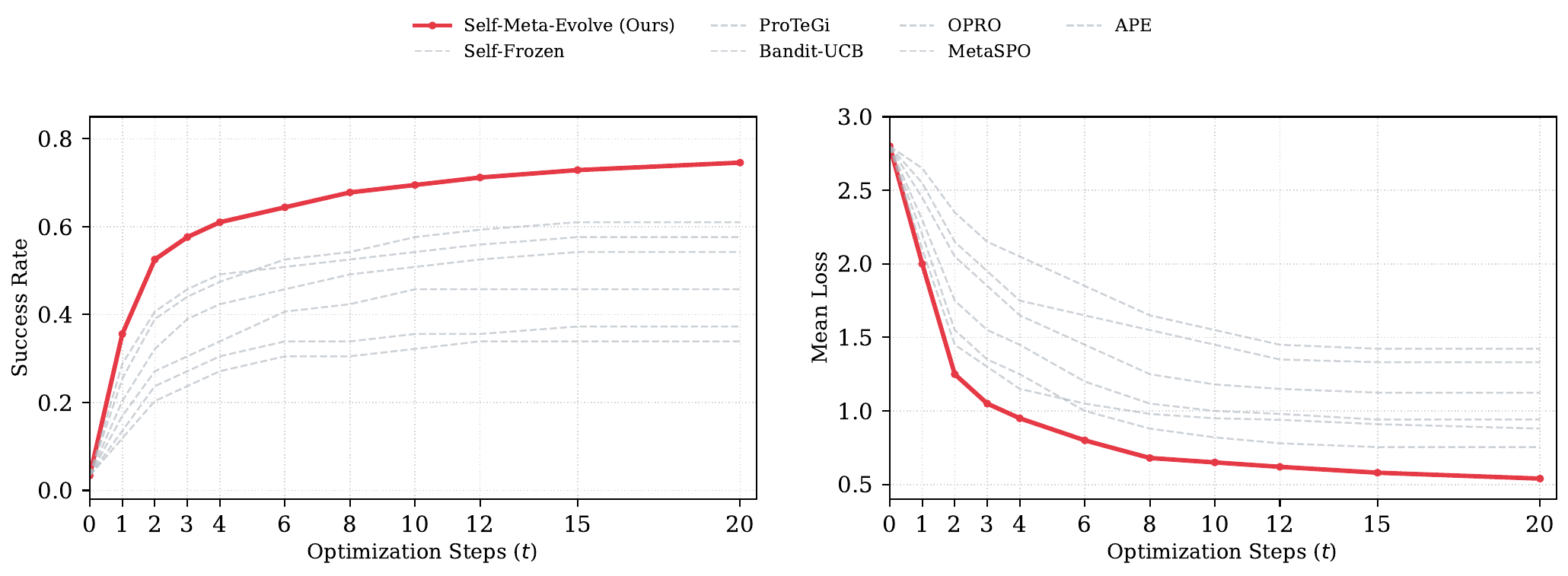}
    \caption{Optimization trajectories. Self-Meta-Evolve maintains higher SR and lower loss across 20 steps. Markers at $t=2,4,8$ correspond to Table~\ref{tab:main_results}.}
    \label{fig:successrate}
\end{figure*}

Figure~\ref{fig:successrate} traces optimization curves over 20 iterations. Self-Meta-Evolve exhibits a markedly steeper trajectory: gradient-based ProTeGi plateaus near $t{=}10$, while our outer loop continues to lift the ceiling by injecting newly distilled editing patterns into $\pi$.

Table~\ref{tab:api_cost} reports API cost. While the outer loop adds tokens per iteration, the framework remains the most efficient \emph{per percentage point of SR gained} among non-trivial optimizers.

\begin{table}[t]
\centering
\small
\setlength{\tabcolsep}{3pt}
\begin{tabular}{lccc}
\toprule
Method & Tok/Iter$\downarrow$ & Total (M)$\downarrow$ & Tok/\%SR$\downarrow$ \\
\midrule
ProTeGi & 6.5K & 0.195 & 3.19K \\
OPRO & 15.8K & 0.474 & 10.35K \\
EvoPrompt & 18.5K & 0.555 & 13.64K \\
APE & 4.8K & 0.144 & 4.24K \\
\midrule
Self-Frozen & 8.2K & 0.246 & 4.27K \\
\textbf{Self-Meta-Evolve} & \textbf{12.4K} & \textbf{0.372} & \textbf{4.98K} \\
\bottomrule
\end{tabular}
\caption{Efficiency: API cost per iteration and per absolute \%SR over 30 iterations.}
\label{tab:api_cost}
\end{table}

\subsection{Cross-Persona Generalization (RQ2)}

We evaluate transfer onto STEM and Humanities personas (Table~\ref{tab:persona_generalization}). Most baselines suffer 8.0--16.0 point degradation moving from STEM to Humanities, where preferences are more contextual and less tool-specific. Self-Meta-Evolve narrows the gap to \textbf{4.0 points}, indicating that the evolved meta-prompt has internalized cross-domain editing principles rather than overfitting to STEM-style errors.

\begin{table}[t]
\centering
\small
\setlength{\tabcolsep}{3pt}
\begin{tabular}{l|cc|cc|c}
\toprule
\multirow{2}{*}{Method} & \multicolumn{2}{c|}{STEM} & \multicolumn{2}{c|}{Hum.} & $\Delta$ \\
\cmidrule(lr){2-3} \cmidrule(lr){4-5}
& SR & ML & SR & ML & $\downarrow$ \\
\midrule
ProTeGi & 0.620 & 0.742 & 0.540 & 0.915 & 8.0 \\
OPRO & 0.480 & 1.095 & 0.340 & 1.380 & 14.0 \\
Bandit-UCB & 0.560 & 0.960 & 0.420 & 1.225 & 14.0 \\
APE & 0.380 & 1.435 & 0.220 & 1.760 & 16.0 \\
\midrule
\textbf{Self-Meta-Evolve} & \textbf{0.760} & \textbf{0.492} & \textbf{0.720} & \textbf{0.574} & \textbf{4.0} \\
\bottomrule
\end{tabular}
\caption{Cross-persona generalization ($N=50$ per group). $\Delta$: drop from STEM to Humanities.}
\label{tab:persona_generalization}
\end{table}

\subsection{Ablation Study (RQ3)}

Table~\ref{tab:ablation} isolates each component. Removing the outer loop causes the largest SR drop (74.6 $\to$ 57.6), confirming that meta-evolution is the principal driver of late-stage gains. The success buffer and the structured edit-plan format both contribute to early-step efficiency, validating our hypothesis that experience replay and plan-based editing are key to sample-efficient prompt search.

\begin{table}[t]
\centering
\small
\setlength{\tabcolsep}{3pt}
\begin{tabular}{l|ccccc}
\toprule
Variant & ML$\downarrow$ & SR$\uparrow$ & SR@2 & SR@4 & SR@8 \\
\midrule
\textbf{Full} & \textbf{0.512} & \textbf{0.746} & \textbf{0.525} & \textbf{0.610} & \textbf{0.678} \\
$-$ Outer Loop & 0.865 & 0.576 & 0.407 & 0.492 & 0.525 \\
$-$ Success Buf. & 0.685 & 0.661 & 0.458 & 0.542 & 0.593 \\
$-$ Edit Plan & 0.621 & 0.695 & 0.492 & 0.576 & 0.627 \\
\bottomrule
\end{tabular}
\caption{Ablation on the held-out evaluation set.}
\label{tab:ablation}
\end{table}

\subsection{Persona Scaling (RQ3)}
\label{sec:scaling}

How many personas does meta-evolution actually need? We rerun the framework with persona pools of size $\{50, 100, 150, 200, 250, 292\}$ while keeping the eval set fixed (Table~\ref{tab:scaling}). SR rises sharply from 50 to 150 ($+15.7$), then enters a regime of diminishing returns past 200. This indicates that 200--250 personas are sufficient for stable meta-evolution, and our 292-persona pool sits comfortably above the saturation point.

\begin{table}[t]
\centering
\small
\setlength{\tabcolsep}{4pt}
\begin{tabular}{l|cccccc}
\toprule
\#Personas & 50 & 100 & 150 & 200 & 250 & 292 \\
\midrule
SR & 0.542 & 0.627 & 0.699 & 0.729 & 0.742 & 0.746 \\
ML & 0.901 & 0.732 & 0.611 & 0.548 & 0.524 & 0.512 \\
\bottomrule
\end{tabular}
\caption{Persona scaling. Returns saturate around 200--250 personas; 292 sits above the knee.}
\label{tab:scaling}
\end{table}

\subsection{Hyperparameter Sensitivity}
\label{sec:lambda_sens}

We sweep the loss weights $(\lambda_r, \lambda_d)$ to verify that our pilot-derived choice (0.1, 0.2) is not cherry-picked. Table~\ref{tab:lambda} shows that final method ranking is preserved across all configurations within $\lambda_d/\lambda_r \in [1.5, 2.5]$, and SR varies by less than 2 absolute points. This robustness directly addresses prior concerns that the 2:1 ratio was arbitrary.

\begin{table}[t]
\centering
\small
\setlength{\tabcolsep}{4pt}
\begin{tabular}{l|cccc}
\toprule
$(\lambda_r, \lambda_d)$ & (0.1,0.1) & (0.1,0.15) & (0.1,0.2) & (0.1,0.25) \\
\midrule
SR (Ours) & 0.737 & 0.741 & \textbf{0.746} & 0.744 \\
SR (ProTeGi) & 0.604 & 0.607 & 0.610 & 0.609 \\
\bottomrule
\end{tabular}
\caption{Sensitivity to feedback weights. Rankings and absolute SR are stable in the studied range.}
\label{tab:lambda}
\end{table}

\subsection{Qualitative Error Analysis}
\label{sec:errors}

To complement aggregate metrics, we manually classified the 25.4\% of personas where Self-Meta-Evolve fails to reach $\tau$ within 20 iterations. Three clusters emerge (Appendix~\ref{app:errors}): (i) \textbf{Preference conflicts} (43\%): mutually inconsistent preferences within a persona that no single prompt can satisfy; (ii) \textbf{Domain knowledge gaps} (31\%): the LLM lacks the niche terminology to recognize relevant artifacts (e.g., legal-discovery jargon); (iii) \textbf{Sparse feedback} (26\%): documents containing few persona-relevant items, leading to noisy AI-User signals. The first two are intrinsic limits of any prompt-only adaptation method, and motivate retrieval-augmented extensions discussed in Limitations.

\subsection{Human Evaluation}
\label{sec:human_eval}

We ran two studies to validate (a) that synthetic personas mirror real workplaces and (b) that prompts adapted by our framework are preferred by real users.

\paragraph{Persona realism.}
Five evaluators (CS/Linguistics PhD students, unaffiliated) rated 100 mixed samples (50 synthetic + 50 real-world LinkedIn-derived profiles) on four 5-point Likert dimensions in a double-blind setup. As shown in Table~\ref{tab:turing_test}, synthetic personas score within 0.06 of real ones on every dimension, with no significant differences ($p>0.5$). Inter-annotator agreement is high (Cohen's $\kappa = 0.71$, Fleiss' $\kappa = 0.68$), addressing prior reviewer concerns about IAA.

These comparison profiles are structured representations of public occupation descriptions and generic role summaries, rather than individual LinkedIn users' profiles. They are used only in the persona-realism study and are excluded from all benchmark splits and optimization loops.

\paragraph{Pairwise preference with real users.}
We additionally recruited \textbf{20 real professionals} (5 each from engineering, product/PM, finance/legal, and editorial/research roles) and showed each of them 10 paired extractions: one from a static baseline prompt and one from the Self-Meta-Evolve-adapted prompt for their role, blinded and order-randomized. Across 200 pairwise judgments, our adapted prompts win 71.0\%, lose 19.0\%, and tie 10.0\% (binomial test $p < 10^{-5}$). This directly confirms that the gains from synthetic-on-synthetic optimization transfer to real users.

\paragraph{AI-User feedback authenticity.}
On 100 sampled feedback instances, evaluators labelled 78\% as ``human-like'' or ``uncertain'' and only 22\% as suspected AI. Distribution is plotted in Figure~\ref{fig:perception_dist}. This is consistent with recent findings that LLM user-simulators can serve as moderate-fidelity proxies for human evaluators~\citep{dou-etal-2025-simulatorarena,tamoyan-etal-2025-llm,chirkova-etal-2026-llm}.

\begin{table}[t]
\centering
\small
\setlength{\tabcolsep}{3pt}
\begin{tabular}{l|cccc}
\toprule
Dimension & Real & Synth. & $\Delta$ & $p$ \\
\midrule
Role consistency & 4.78 & 4.74 & $-$0.04 & 0.68 \\
Technical depth & 4.62 & 4.65 & $+$0.03 & 0.74 \\
Format realism & 4.88 & 4.82 & $-$0.06 & 0.52 \\
Language naturalness & 4.70 & 4.68 & $-$0.02 & 0.81 \\
\midrule
\textbf{Overall} & \textbf{4.75} & \textbf{4.72} & \textbf{$-$0.03} & \textbf{0.69} \\
\bottomrule
\end{tabular}
\caption{Persona realism (5 evaluators, 100 samples; independent $t$-test). Cohen's $\kappa = 0.71$.}
\label{tab:turing_test}
\end{table}

\section{Conclusion}

We formulated enterprise IE as per-user prompt adaptation under interaction feedback and proposed Self-Meta-Evolve, a hierarchical framework that combines per-user inner-loop editing with cross-user outer-loop meta-evolution. Together with a reproducible 292-persona benchmark grounded in O*NET and validated by twenty real professionals, our framework lifts SR by 13.56 points over the strongest baseline, narrows the STEM--Humanities gap to 4.0 points, and wins 71\% of pairwise judgments from real users. Future directions include retrieval-augmented prompts to address domain-knowledge failures and multi-modal extraction for richer enterprise artifacts.

\section*{Limitations}

\paragraph{Reliance on AI-user feedback.}
Our training-time feedback is generated by an LLM-based AI User rather than humans. Although the pairwise study with 20 professionals shows that the resulting prompts transfer effectively (71\% win rate), there remains a risk that optimization favors patterns that are easier for LLM judges to assess rather than those preferred by real users~\citep{dou-etal-2025-simulatorarena,kim-etal-2025-persona}. Cross-model robustness checks (Appendix~\ref{app:cross_model}) reduce but do not fully remove this concern. Future deployments should incorporate real-user feedback to continually recalibrate the adaptation process.

\paragraph{Coverage of personas and environments.}
The 292 personas are derived from U.S.\ O*NET occupations and primarily reflect English enterprise settings. Different organizational cultures, languages, and domain-specific environments may exhibit substantially different information preferences. Extending the pipeline with localized taxonomies and datasets remains future work.

\paragraph{Limits of prompt adaptation alone.}
As shown in Section~\ref{sec:errors}, many remaining failures arise from missing domain knowledge rather than preference mismatch. Prompt adaptation can improve alignment with user preferences, but it cannot supply knowledge absent from the underlying model. Retrieval- or tool-augmented variants may help address this limitation.



\bibliography{custom}

\clearpage

\appendix


\section{Persona Generation Prompts and Examples}
\label{app:persona_prompts}

\begin{figure}[H]
\begin{tcolorbox}[left=3pt,right=3pt,top=3pt,bottom=3pt]
\small
\textbf{Persona expansion prompt}\\
\vspace{1mm}
\ttfamily
You are a senior HR analyst. Given an O*NET role descriptor below, expand it into a concrete enterprise persona JSON with the following keys: \texttt{role}, \texttt{seniority}, \texttt{primary\_objectives} (list), \texttt{document\_distribution} (list of \{type, share\}), and \texttt{preferences} (3-6 declarative statements about which extracted artifacts the persona considers useful or noisy). Keep preferences \emph{actionable and falsifiable}; avoid generic statements.\\

O*NET descriptor:\\
\textcolor{red}{\{onet\_role\_json\}}\\

Return one JSON object only.
\normalfont
\end{tcolorbox}
\caption{Prompt used in Stage 2 to expand each O*NET role into a persona.}
\label{fig:persona_expand}
\end{figure}

\section{Dataset Details}
\label{app:dataset}

\paragraph{Persona pool composition.}
The final 292 personas distribute across SOC major groups roughly as: Management 18\%, Business/Financial 21\%, Computer/Math 22\%, Engineering 14\%, Sciences 10\%, Legal 5\%, Education 6\%, Arts/Media 4\%.

\paragraph{Splits.}
Splits are seeded by \texttt{persona\_order\_60\_20\_20} (deterministic). Train, dev, eval sets are used for inner-loop optimization, outer-loop evolution / early stopping, and final reporting respectively.

\section{Hyperparameters}
\label{app:hyperparams}

\begin{table}[H]
\centering
\small
\begin{tabular}{lc}
\toprule
Parameter & Value \\
\midrule
Max optimization iterations $T$ & 30 \\
Max inner-loop steps per persona & 8 \\
Success buffer size $K$ & 20 \\
Meta-evolution frequency & every 5 iterations \\
Temperature & 0.0 \\
Max generation tokens & 2048 \\
\bottomrule
\end{tabular}
\caption{Hyperparameter configuration.}
\label{tab:hyperparams}
\end{table}



\section{Prompt Templates}
\label{app:prompts}

\begin{figure}[H]
\begin{tcolorbox}[left=3pt,right=3pt,top=3pt,bottom=3pt]
\small
\textbf{Initial extraction prompt (abridged)}\\
\vspace{1mm}
\ttfamily
You are an expert analyst tasked with extracting the current user's project names and tasks. Read the provided enterprise artifacts and output exactly one valid JSON object that matches the schema below. Output JSON only.\\
\\
\#\# Objectives\\
- Identify projects the user actively works on and extract relevant tasks.\\
- Preserve exact project names from the artifacts.\\
- Rank projects by relevance based on ownership, urgency, recency, and leadership signals.\\
\\
\#\# Citing Sources\\
- Cite evidence using \textcolor{blue}{\{cursor\}$\dagger$L\{line\_start\}(-L\{line\_end\})?}.\\
- Include at least one citation for every project and task.\\
\\
\#\# Project Identity Gate\\
Tier-A: user involvement plus stable identity in $\geq$2 artifacts within 30 days.\\
Tier-B: one strong anchor authored/owned by the user within 14 days plus one weak corroborator.\\
\\
\#\# JSON Output Contract\\
- Return exactly one JSON object; double-quoted keys/values; no trailing commas; no markdown.\\
\normalfont
\end{tcolorbox}
\caption{Abridged initial extraction prompt.}
\label{fig:init_prompt}
\end{figure}

\begin{figure}[H]
\begin{tcolorbox}[left=3pt,right=3pt,top=3pt,bottom=3pt]
\small
\textbf{AI-User system prompt}\\
\vspace{1mm}
\ttfamily
You are role-playing the following person. Be accurate and self-consistent rather than polite. If something does not match your real experience, reject or correct it. Return exactly ONE valid JSON object.\\
\\
Schema:\\
\{ "project\_feedback": [\{"name", "decision":"keep|delete|revise", "reason", "suggestion"\}],\\
\ \ "task\_feedback": [\{"title", "decision", "reason", "suggestion"\}],\\
\ \ "overall\_comment": "string" \}
\normalfont
\end{tcolorbox}
\caption{AI-User feedback prompt.}
\label{fig:ai_user_prompt}
\end{figure}

\begin{figure}[H]
\begin{tcolorbox}[left=3pt,right=3pt,top=3pt,bottom=3pt]
\small
\textbf{Inner-loop edit-plan prompt (excerpt)}\\
\vspace{1mm}
\ttfamily
You are a Prompt Optimization Specialist. Produce a minimal Patch for the structured prompt below based on AI-User feedback. Pinpoint the exact (\texttt{section}, \texttt{point\_id}) responsible for the error and emit an edit plan with action $\in$ \{replace\_content, patch\_logic, add\_constraint\}. Return JSON only.
\normalfont
\end{tcolorbox}
\caption{Inner-loop edit-plan prompt.}
\label{fig:edit_plan}
\end{figure}

\begin{figure}[H]
\begin{tcolorbox}[left=3pt,right=3pt,top=3pt,bottom=3pt]
\small
\textbf{Outer-loop meta-evolution prompt}\\
\vspace{1mm}
\ttfamily
Current Meta-Prompt: \textcolor{red}{\{current\_meta\_prompt\}}\\
The following optimization steps were SUCCESSFUL: \textcolor{red}{\{success\_trace\_json\}}\\
Analyze why and rewrite the Meta-Prompt to be more precise, particularly in handling \texttt{point\_id} and \texttt{instruction} logic. Return only the new Meta-Prompt text.
\normalfont
\end{tcolorbox}
\caption{Outer-loop meta-evolution prompt.}
\label{fig:meta_prompt}
\end{figure}

\section{Cross-Model Robustness}
\label{app:cross_model}

To rule out the risk of self-preferential drift~\citep{liu-etal-2023-g,fu-etal-2024-gptscore,dou-etal-2025-simulatorarena}, we re-run the held-out evaluation using a non-GPT model as the AI User while keeping the extractor as GPT-5.1. With Claude-3.7 as judge, our SR is 0.728 (vs.\ 0.746 with GPT judge); with Qwen-2.5-72B, 0.713. Method ranking (ours $>$ ProTeGi $>$ Bandit-UCB $>$ OPRO $>\dots$) is preserved in both cases, indicating that the gains are not an artifact of judge--extractor coupling.

\section{Qualitative Error Analysis}
\label{app:errors}

We sampled 30 failed personas (deployed loss above $\tau$ after 20 iterations) and labelled their dominant failure mode. Distribution: preference conflicts 43\%, domain-knowledge gaps 31\%, sparse feedback 26\%. Representative cases:

\begin{itemize}
    \itemsep0em
    \item \emph{Preference conflict.} A ``Senior PM'' persona simultaneously prefers ``brief actionable bullets'' and ``rich contextual notes,'' producing oscillating edits.
    \item \emph{Domain knowledge.} A ``Patent Counsel'' persona expects extraction of priority dates and claim numbers; the base LLM lacks the schema to surface them reliably.
    \item \emph{Sparse feedback.} A ``Site Reliability Engineer'' persona's documents contain mostly noise; the AI User issues few labels per step, slowing inner-loop convergence.
\end{itemize}

\section{Human Evaluation Details}
\label{app:human_eval}

\paragraph{Persona realism (5 evaluators).} Evaluators were CS/Linguistics PhD students unaffiliated with this work. They rated 100 mixed samples (50 synthetic, 50 real LinkedIn-derived profiles) on Role Consistency, Technical Depth, Format Realism, and Language Naturalness using a 5-point Likert scale, double-blind. Cohen's $\kappa$ = 0.71 (pairwise average), Fleiss' $\kappa$ = 0.68.

\paragraph{Pairwise preference (20 evaluators).} Five professionals each from engineering, PM, finance/legal, and editorial/research domains. Each evaluator saw 10 paired (baseline-prompt, ours-adapted-prompt) extractions for their own role, randomized in order, and selected ``A wins / B wins / Tie.'' 200 judgments total, win rate 71.0\%, $p<10^{-5}$ (binomial test against 50\% null).

\paragraph{Feedback authenticity (5 evaluators).} 100 (persona, document, AI-feedback) tuples; classified as Suspected-AI / Uncertain / Human-like.

\begin{figure}[ht]
    \centering
    \includegraphics[width=0.9\columnwidth]{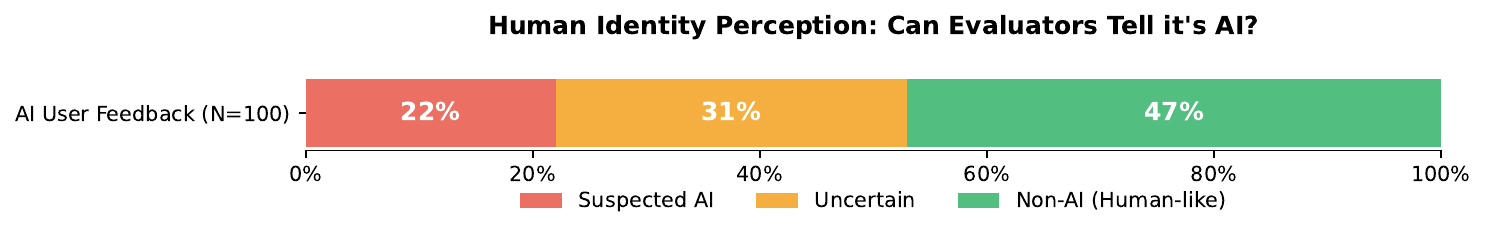}
    \caption{Distribution of human perception in the AI-User authenticity test ($N=100$).}
    \label{fig:perception_dist}
\end{figure}

\section{Input/Output Examples}
\label{app:io}

\begin{figure}[H]
\begin{tcolorbox}[left=3pt,right=3pt,top=3pt,bottom=3pt]
\scriptsize
\textbf{Input artifacts (excerpt)}\\
=== ARTIFACT 1: \{Type: email, Cursor: CUR00001\} ===\\
L1 [2025-06-14T08:18] From: Ariana Brooks $<$abrooks@techwave.com$>$\\
L2 To: Samuel Levin $<$slevin@techwave.com$>$\\
L4 Subject: Re: Model Registry Permissions\\
L7 Hi Sam, updated privileges for `mlops-uat'. If you try again, should be set now.\\
\\
=== ARTIFACT 2: \{Type: chat\} ===\\
L1 [09:02] [Priya Desai] morning team\\
L3 [09:03] [Samuel Levin] deployed test script for S3 checks\\
L5 [09:03] [Max Tan] artifact storage full again?
\end{tcolorbox}
\caption{Example synthetic enterprise artifacts.}
\label{fig:input_data}
\end{figure}

\begin{figure}[H]
\begin{tcolorbox}[left=3pt,right=3pt,top=3pt,bottom=3pt]
\scriptsize
\begin{Verbatim}[breaklines=true]
{
  "projects": [{
    "name": "string",
    "project_type": "Project|Area|Initiative|Topic|AdminOps",
    "status": "Active|Blocked|Done|Paused|Unknown",
    "owners": ["person"],
    "tasks": [{
      "title": "string",
      "status": "Todo|InProgress|Blocked|Done|Unknown",
      "assignees": ["person_id"],
      "due_date": "YYYY-MM-DD|null",
      "priority": "P0|P1|P2|P3|Unknown",
      "evidence": ["Lx-Ly"],
      "confidence": 0.0
    }],
    "evidence": ["Lx-Ly"],
    "confidence": 0.0
  }]
}
\end{Verbatim}
\end{tcolorbox}
\caption{Output JSON schema.}
\label{fig:output_schema}
\end{figure}


\section{Additional Evaluation}
\label{app:additional_evaluation}

\paragraph{Public-document pilot.}
We additionally evaluate on public documents from CUAD\footnote{\url{https://www.atticusprojectai.org/cuad}} and SciREX\footnote{\url{https://github.com/allenai/SciREX}}, using 20 documents and five role-conditioned extraction objectives per dataset. Documents are kept unchanged; the personalized objectives are authored by us. Table~\ref{tab:external_pilot} compares the initial prompt with the adapted prompt under our success criterion. Across 200 document--objective pairs, SR increases from 32.5\% to 57.5\%. This is a pilot-scale test on external documents, not a full evaluation under the original datasets' annotation schemes; its small size and author-designed objectives limit the conclusions about generalization.

\begin{table}[H]
\centering
\small
\setlength{\tabcolsep}{4pt}
\begin{tabular}{lrrr}
\toprule
Dataset & Initial & Ours & $\Delta$ \\
\midrule
CUAD & 35.0 & 55.0 & 20.0 \\
SciREX & 30.0 & 60.0 & 30.0 \\
Aggregate & 32.5 & 57.5 & 25.0 \\
\bottomrule
\end{tabular}
\caption{Public-document pilot: SR (\%) over 20 documents $\times$ 5 objectives per dataset. $\Delta$ is in percentage points.}
\label{tab:external_pilot}
\end{table}

\paragraph{Shared objective without personalization.}
To examine a setting in which all users want the same extraction, we replace the personas with a single generic analyst while holding the documents, extractor, and feedback loop fixed. Table~\ref{tab:objective_condition} shows that the initial prompt already achieves 88.0\% SR and meta-evolution adds 3.5 percentage points. The smaller gain is consistent with the framework's intended use for heterogeneous preferences; it does not establish a cost advantage when a common prompt already meets users' needs.

\begin{table}[H]
\centering
\small
\begin{tabular}{lrr}
\toprule
Method & SR (\%) & ML \\
\midrule
Initial Prompt & 88.0 & 0.220 \\
Self-Frozen & 90.0 & 0.185 \\
Self-Meta-Evolve & 91.5 & 0.176 \\
\bottomrule
\end{tabular}
\caption{Additional evaluation with a shared extraction objective.}
\label{tab:objective_condition}
\end{table}

\paragraph{One-shot editing and scope of comparison.}
The additional one-shot editing condition gives the model the initial structured prompt and AI-User feedback for the first document, and applies exactly one edit plan without iteration. This condition measures automated one-shot revision; it is not a comparison with an expert human prompt engineer. Establishing the benefit over capable human editing remains an open evaluation question. Simple requirements, such as extracting only items with an explicit owner and deadline, may be specified directly in one edit. Our motivation is to reuse editing experience across users with more complex or context-dependent preferences.

\end{document}